\documentclass[a4paper]{svproc}
\usepackage[numbers, sectionbib]{natbib}
\usepackage{multicol}
\usepackage[colorlinks,bookmarksopen,bookmarksnumbered,citecolor=red,urlcolor=red, bookmarks=true]{hyperref}
\usepackage{color}
\usepackage{graphicx,amsmath,amssymb,bm,xcolor}
\usepackage{subcaption} 
\usepackage{multirow} 
\usepackage[utf8]{inputenc}
\usepackage{breqn}
\usepackage[title]{appendix}
\usepackage[ruled,norelsize]{algorithm2e}
\usepackage{mathtools} 
\usepackage[noabbrev]{cleveref}

\usepackage{todonotes}
\usepackage{dsfont}
\ifdefined\N                                                                
\renewcommand{\N}{\mathds{N}}                                                
\else
\newcommand{\N}{\mathds{N}}
\fi
\newcommand{\R}{\mathds{R}}                                                 
\ifdefined\C
\renewcommand{\C}{\mathds{C}}                                             
\else
\newcommand{\C}{\mathds{C}}
\fi

\def\argmin{\mathop{\sf arg\,min}}                                          
\newcommand{\abs}[1]{\left|#1\right|}

\newcommand{\id}{\boldsymbol{I}}                                                
\newcommand{\diag}{\operatorname{diag}}                                     
\newcommand{\blkdiag}{\operatorname{blkdiag}}                               
\newcommand{\trace}[1]{\operatorname{Tr}[#1]}                                      

\newcommand{\matb}[1]{ 														
	\begin{bmatrix}
		#1
	\end{bmatrix}
}
\newcommand{\trsp}{{\scriptscriptstyle\top}}									

\newcommand{\E}{\mathds{E}}                                                 
\newcommand{\normal}{\mathcal{N}}                                           

\newcommand{\nd}[1]{\bm{#1}}
\newcommand{\st}{\mbox{s.t.}}
\newcommand{\norm}[1]{\lVert #1 \rVert}

\newcommand{\Log}{\text{Log}}

\newcommand{\Phii}{\bm{\Phi}}
\newcommand{\Phix}{{\Phii_x}}
\newcommand{\Phiu}{{\Phii_u}}
\newcommand{\SFpair}{\{\Phix, \Phiu\}}

\newcommand{\deltaxt}{\nd x_t - \bm{\hat{x}}_t}

\newcommand{\deltaut}{\nd u_t - \bm{\hat{u}}_t}

\newcommand{\Su}{\bm{S_u}}
\newcommand{\Sx}{\bm{S_x}}

\usepackage{url}

\begin{document}
\mainmatter              
\title{Reactive Anticipatory Robot Skills with Memory}
\titlerunning{Reactive Anticipatory Robot Skills with Memory}  
%
\author{Hakan Girgin  \and Julius Jankowski \and
Sylvain Calinon }
\authorrunning{Hakan Girgin et al.} 
%
\tocauthor{Hakan Girgin, Julius Jankowski and Sylvain Calinon}
\institute{Idiap Research Institute, Martigny 1920, Switzerland,\\
\email{\{hakan.girgin, julius.jankowski, sylvain.calinon\}@idiap.ch}
}

\maketitle              

\begin{abstract}
Optimal control in robotics has been increasingly popular in recent years and has been applied in many applications involving complex dynamical systems. Closed-loop optimal control strategies include model predictive control (MPC) and time-varying linear controllers optimized through iLQR. However, such feedback controllers rely on the information of the current state, limiting the range of robotic applications where the robot needs to remember what it has done before to act and plan accordingly. The recently proposed system level synthesis (SLS) framework circumvents this limitation via a richer controller structure with memory. In this work, we propose to optimally design reactive anticipatory robot skills with memory by extending SLS to tracking problems involving nonlinear systems and nonquadratic cost functions. We showcase our method with two scenarios exploiting task precisions and object affordances in pick-and-place tasks in a simulated and a real environment with a 7-axis Franka Emika robot.
\end{abstract}
\section{Introduction}
Optimal control can be found in a great variety of applications from economics \cite{kamien2012dynamic} to engineering problems such as energy management \cite{bianchi2007wind} and robotics \cite{Diehl2006,OC2}. It consists of determining optimal actions in problems following a known forward model that describe state changes in time when actions are applied.

In robotics, optimal control has been applied in many applications such as biped walking generation \cite{Kajita2003,Caron2016}, centroidal dynamics
trajectory \cite{Ponton2018,Winkler2018} and whole-body motion planning \cite{Budhiraja2019}. These works often consider the solution of optimal trajectories of states and actions as a plan over a horizon, which is then tracked with lower level feedback control mechanisms. As the forward models that define these actions are not perfect representations of the real physical movements, feedback control is designed to achieve the planned motion even in the presence of noise, delays and unpredictable perturbations in the environment. 

\begin{figure}[t!]
	\begin{subfigure}{0.38\columnwidth}
		\centering
		\includegraphics[width=0.8\columnwidth]{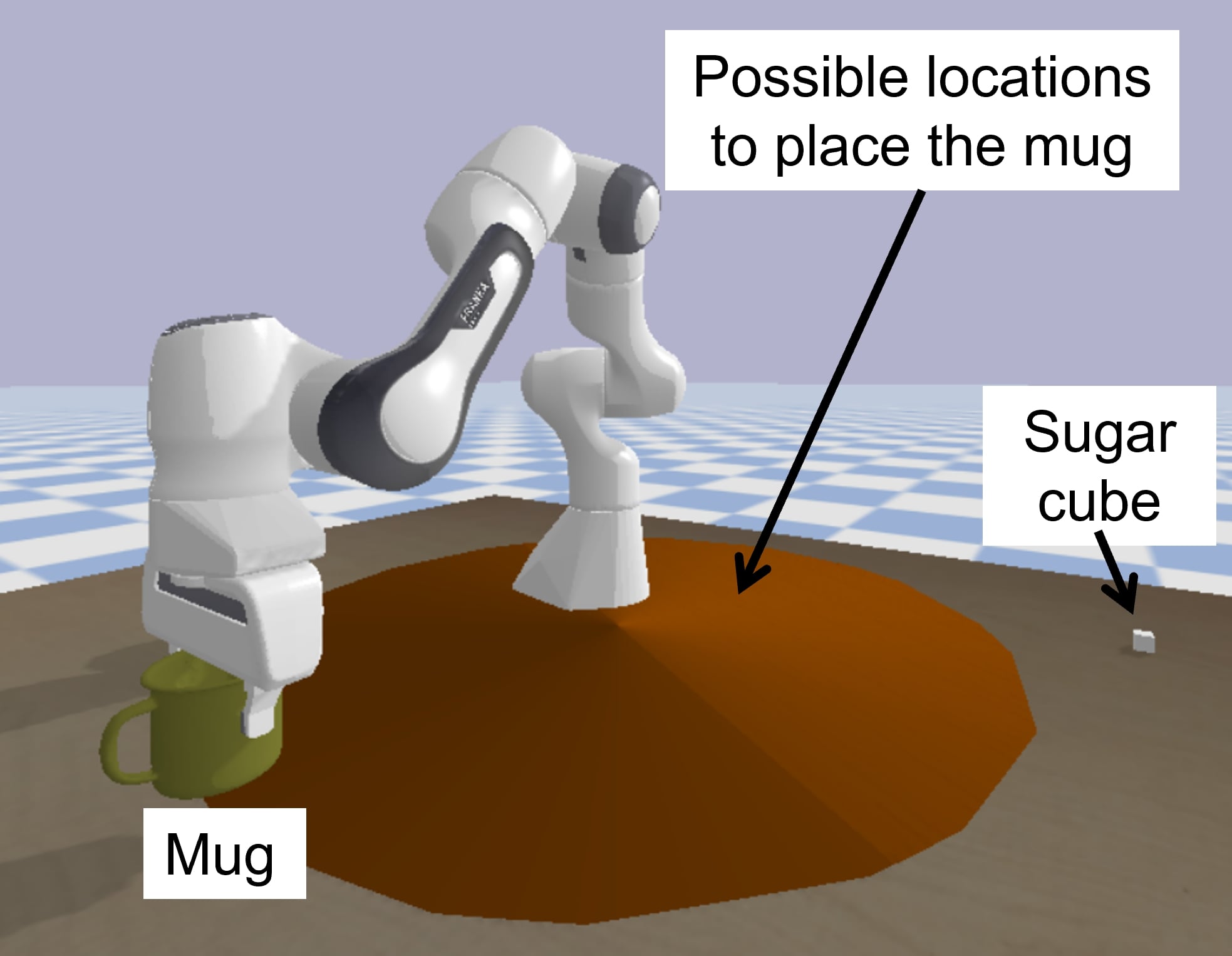}
		\caption{Mug-sugar cube scenario}
		\label{fig:setup1}
	\end{subfigure}\hfill
	\begin{subfigure}{0.6\columnwidth}
		\centering
		\includegraphics[width=0.98\columnwidth]{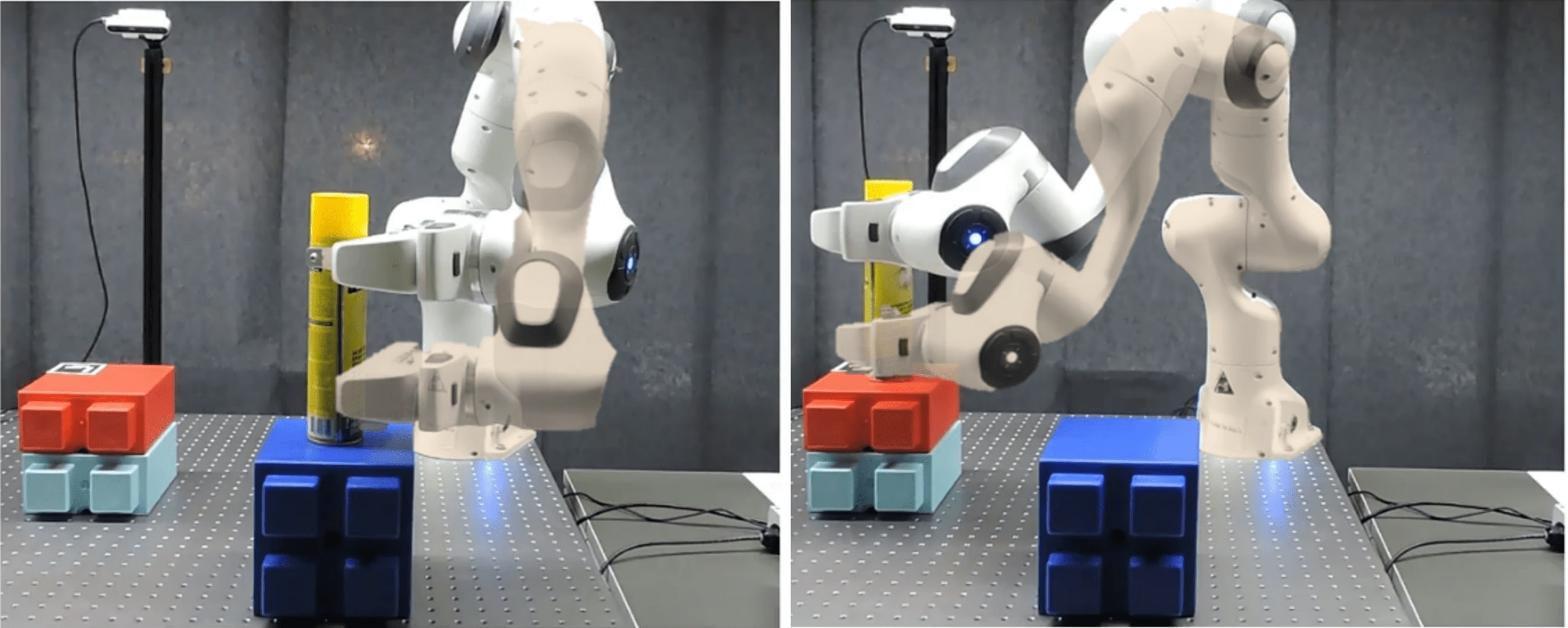}
		\caption{Pick-and-place task}
		\label{fig:setup2}
	\end{subfigure}
	\caption{(a) The robot decides where to place the object according to the initial position and with the anticipation to pick up the sugar cube and put it inside the cup.  (b) Pick-and-place task with the Franka Emika robot deciding autonomously to grasp the object from different parts, as an adaptation to different initial configurations. The memory feedback property allows the robot to remember where it placed the mug and grasped the object, so that it can place it accordingly.\vspace{-0.5cm}}
	\label{fig:setup}
\end{figure}

Among others, model predictive control (MPC) \cite{MPC} is a powerful feedback mechanism in optimal control, that became a key methodology to control complex dynamical systems such as humanoids \cite{humanoid}. Linear quadratic tracking (LQT) \cite{lqt} and iterative linear quadratic regulator (iLQR) \cite{ilqr} and their variants are simple solvers that are increasingly used in MPC frameworks for controlling high dimensional robotic systems such as humanoids and manipulators \cite{kleff21,quadruped,humanoid,legged}. In particular, dynamic programming solutions for these methods are often computationally less intensive than other trajectory optimization methods, with the additional benefit of outputting a full state feedback controller with gains over the horizon. The feedback controller structure in iLQR enables the feedback gains to react to the current state of the system while anticipating the future states. The current state contains implicitly the information from past states; however, the controller cannot directly react on this past information explicitly, resulting in mediocre performance in applications where the robot needs to remember what it has done before to react and replan. For example, a robot grasping an object from its different parts needs to remember where it grasped the object to place it without collision as illustrated in \Cref{fig:setup2}. One solution is to augment the state-space with the past states and solve the problem in the standard way. However, it would result in a much higher dimensional problem which may be impractical for robotic applications.

Recently, System Level Synthesis (SLS) \cite{anderson19} was proposed as an optimal controller reacting not only to the current state but also to the past states of the system through closed-loop mappings optimization. Such a controller with memory could be exploited in tasks requiring correlations between states at different timesteps, as opposed to a controller without memory such as a LQR controller. However, the SLS framework has a couple of limitations preventing it from being used in robotics applications requiring a memory of states with sparse cost functions and nonlinear dynamics.

In this work, we investigate how the optimal control framework can produce anticipatory and reactive robot skills with a memory of the states to produce smart behaviors that can exploit task precisions and object affordances. We propose to achieve our goal with the following extensions to the SLS framework: 1) by adding a feedforward part to the controller enabling trajectory and viapoint tracking, 2) by extending the approach to nonquadratic costs and nonlinear systems using a Newton method optimization; and 3) by providing fast adaptation and replanning strategies.

The paper is organized as follows. First, \Cref{sec:related_work} presents the related work in controllers with memory and SLS controllers in robotics. In \Cref{sec:background}, we introduce the SLS framework as background. Then, in \Cref{sec:methods}, we show our proposed extensions over SLS that enable its application in robotics and the proposed adaptation strategies when the robot encounters new situations. In \Cref{sec:exp}, we showcase our approach with two scenarios exploiting task precisions and object affordances in pick-and-place tasks in a simulated and a real environment with a 7-axis Franka Emika robot. Finally, we conclude the work and discuss potential impacts of the work in \Cref{sec:conclusion}.

\section{Related Work}
\label{sec:related_work}
Feedback controllers that can act on the history of states are investigated in robotics mainly via reinforcement learning algorithms by changing the structure of the policy, especially with recurrent neural networks (RNN) \cite{Deisenroth13}. In \cite{Siekmann20}, the authors propose to learn stable bipedal locomotion with an RNN policy, where the memory serves as a model of the physical parameters of the task. In \cite{Zhang16}, a deep learning architecture for learning a policy that can remember some important information from the past observations is developed by augmenting the policy state with a continuous memory state. The authors showed that their guided policy search algorithm could solve a peg sorting task with a manipulator which needs to remember the target hole position given at the beginning of the training and a plate and bottle placing task where the robot needs to remember which object it is holding to determine the orientation. 

System level synthesis (SLS) has been developed in a collection of works that are summarized in \cite{anderson19}. It provides a novel perspective on robust optimal control design by optimizing over the closed-loop mappings of the system, instead of directly optimizing the controller. The authors showed how SLS can be exploited in the domain of large-scale distributed optimal control and robust optimal control.

In \cite{dean19}, SLS has been exploited to learn unknown dynamics for safe control with LQR, including robust safety guarantees in the state and input constraints. In \cite{dean20}, perception based SLS controllers are designed for autonomous control of vehicles learning linear time invariant dynamical systems from image data. In \cite{laura20}, robust perception-based SLS controller is applied for the safe control of a quadrotor. The work in \cite{dimitar20} gives necessary and sufficient conditions for the existence of SLS controller for nonlinear systems and \cite{jing20} exploits these ideas by designing a nonlinear controller for a constrained LQR problem by blending several linear controllers.

\section{Background}
\label{sec:background}
This section follows \cite{anderson19} to present the SLS framework for regulation problems. A linear-time-varying system can be written as $\bm{x}_{t+1} = \bm{A}_t\bm{x}_t + \bm{B}_t\bm{u}_t + \bm{w}_t, \forall t{=}\{0,\ldots,T\}$, where $\bm{x}_t \in \R^m$ is the state, $\bm{u}_t \in \R^n$ is the control input, and $\bm{w}_t \in \R^m$ is an exogenous disturbance term. By stacking these vectors for each time-step $t$, we define $\bm{x} {=} \begin{bmatrix} \bm{x}_0^\trsp & \bm{x}_1^\trsp & \hdots & \bm{x}_{T}^\trsp \end{bmatrix}^\trsp$, $\bm{u} {=}\begin{bmatrix} \bm{u}_0^\trsp & \bm{u}_1^\trsp & \hdots & \bm{u}_{T}^\trsp \end{bmatrix}^\trsp$ and $\bm{w} {=} \begin{bmatrix} \bm{x}_0^\trsp  & \bm{w}_0^\trsp & \bm{w}_1^\trsp & \hdots & \bm{w}_{T-1}^\trsp\end{bmatrix}^\trsp$, which allows us to express the dynamics as
\begin{align}
	\bm{x} &= \bm{Z}\bm{A}_d\bm{x} + \bm{Z}\bm{B}_d\bm{u} + \bm{w},
	\label{eq:stacked_dynamics}
\end{align}
where $\bm{Z}$ is a delaying operator with identity matrices along its first block sub-diagonal and zeros elsewhere, $\bm{A}_d{=} \blkdiag\left(\bm{A}_0,\bm{A}_1,\dots,\bm{A}_{T}\right)$ and $\bm{B}_d{=} \blkdiag( \linebreak\bm{B}_0,\bm{B}_1,\dots,\bm{B}_{T})$. In this work, we consider the stacked disturbance as $\nd w{\sim}\normal(\nd \mu_{w},\linebreak \nd \Sigma_{w})$, where $\nd \mu_w{=}[\nd \mu_{x_0}^\trsp, \nd 0^\trsp]^\trsp$ and $\nd \Sigma_{w}{=}\mathrm{blkdiag}(\nd \Sigma_{x_0},\nd \Sigma_{\text{noise}})$. 

We assume a controller of the form $\nd u{=}\nd K \nd x$, where $\nd K$ is a lower block triangular matrix. Note that this controller is more advanced than the controller found by linear quadratic regulator (LQR). In fact, the former includes the latter at its block-diagonal elements while the off-block-diagonal elements act on the history of the states. Inserting the controller definition into \eqref{eq:stacked_dynamics}, we get $\nd x{=}\bm{Z}\bm{A}_d\bm{x} + \bm{Z}\bm{B}_d\bm{K}\bm{x} + \bm{w}$, which results in the closed-loop responses
\begin{align}
	\bm{x}&=\Phix\bm{w}, \quad \Phix = {\Big(\id -\bm{Z}(\bm{A}_d+\bm{B}_d\bm{K})\Big)}^{-1}, \label{eq:resp_x}\\
	\bm{u}&=\Phiu\bm{w}, \quad \Phiu = \bm{K}\Big(\id - \bm{Z}(\bm{A}_d+\bm{B}_d\bm{K})\Big)^{-1}, \label{eq:resp_y}
\end{align} 
where $\SFpair$ describe the closed loop system responses from the exogenous disturbance $\nd w$ to the state $\nd x$ and control input $\nd u$, respectively. SLS optimizes directly over these system responses, instead of the controller map $\nd K$ as was done by the dynamic programming solutions of LQR. As the controller $\nd K$ is a block lower triangular matrix, these maps are also lower block triangular matrices.

By inserting \eqref{eq:resp_x} and \eqref{eq:resp_y} into the dynamics equation \eqref{eq:stacked_dynamics}, we obtain
\begin{align}
	\Phix\bm{w} &=\bm{Z}\bm{A}_d\Phix\bm{w}+\bm{Z}\bm{B}_d\Phiu\bm{w} + \bm{w}, \nonumber \\
	\implies \Phix &= \bm{Z}\bm{A}_d\Phix + \bm{Z}\bm{B}_d\Phiu + \id = \nd S_{\nd x} + \nd S_{\nd u} \Phiu,
	\label{eq:resp_dynamics}
\end{align}
where the implication is because we want this pair of system responses to remain independent of the noise in the system for every initial state, and $\nd S_{\nd x} {=} (\id - \nd Z \nd A_d)^{-1}$, $\nd S_{\nd u} {=} \nd S_{\nd x} \nd Z \nd B_d$.
The objective of SLS is to optimize directly over these lower block triangular system responses $\SFpair$ so that there exists a linear controller $\bm{K}$ such that $\bm{K}{=}\Phiu \nd \Phi^{-1}_x$ and \eqref{eq:resp_dynamics} holds as outlined by the Theorem 2.1 of \cite{anderson19}.

\textbf{Linear quadratic regulator:} We consider now a linear quadratic regulator problem with random noise in the process and with a random initial state as
\begin{equation}
	\begin{array}{rl}
		\min_{\nd x_t,\nd u_t} & \sum_{t=0}^{T} \E[\nd x_t^\trsp \nd Q_t \nd x_t + \nd u_t^\trsp \nd R_t \nd u_t]\\
		\st & \nd x_{t+1} = \nd A \nd x_t + \nd B \nd u_t + \nd w_t.
	\end{array}
	\label{eq:lqr_toep}
\end{equation}
Using the stacked vectors $\nd x$, $\nd u$ and $\nd w$ and the block diagonal matrices $\nd Q=\mathrm{blkdiag}(\nd Q_0,\nd Q_1, \dots, \nd Q_{T})$ and $\nd R{=}\mathrm{blkdiag}(\nd R_0,\nd R_1, \dots, \nd R_{T})$, this problem can be recast as optimization over system responses as
\begin{equation}
	\begin{array}{rl}
		\min_{\Phix,\Phiu} & \E[\norm{\Phix \nd w}_{\nd Q}^{2} +  \norm{\Phiu \nd w}_{\nd R}^{2}]  \\ 
		\st &  \Phix =\nd S_{x}+\nd S_{u} \Phiu,  \\
		& \Phix, \Phiu \in \mathcal{L} 
	\end{array}
	\label{eq:lqr_compact}
\end{equation}
where $\mathcal{L}$ represents the space of lower block triangular matrices. Solving \eqref{eq:lqr_compact}, we obtain a feedback controller achieving the desired system responses. One can show that by removing the expectation and treating the cost function as random valued, one can solve the problem independently of the value of $\nd w$. Therefore, in the remainder of this paper, we take the cost functions as random valued cost functions instead of their expectations.

Note that because of the constraint to lie on the space $\mathcal{L}$, this problem cannot be solved directly as we would solve a simple LQR problem. Instead, we make use of the column separability of the objective function and the equality constraint representing the dynamics model to separate the problem into $T$ independent subproblems containing only the nonzero part of each block column of the system responses as in \cite{anderson19}.

The controller found by solving \eqref{eq:lqr_compact} can be applied directly to problems with linear forward models and quadratic costs, if the task is to regulate the state to a set-point or to a reference trajectory, which is already dynamically feasible by the plant. In fact, in these cases, one can transform the forward state dynamics to forward error dynamics and still exploit SLS to solve such tasks. However, in most of the robotics applications, the reference trajectory is composed of sparse viapoints and/or the system is nonlinear and/or the cost is nonquadratic. In the next section, we show how to alleviate these problems as part of our contributions.

\section{Methods}
\label{sec:methods}
In this section,  we show that we can solve SLS problems for tracking analytically and how to solve them for nonlinear systems and nonquadratic costs. In Section \ref{subsec:LQT}, we extend the SLS method to \emph{extended system level synthesis} (eSLS) by adding a feedforward part that describes the desired states and the desired control commands. Then, in Section \ref{subsec:isls}, we exploit eSLS to solve the problems with nonquadratic cost and nonlinear dynamics, resulting in an \emph{iterative system level synthesis} (iSLS) approach, which greatly extends the domain of applications of the standard SLS.
\subsection{Extended system level synthesis (eSLS)}
\label{subsec:LQT}
The closed loop responses in \eqref{eq:resp_x} and \eqref{eq:resp_y} cannot represent the closed-loop dynamics well when the problem is to track a desired state ${\bm{x}_{(d,t)}}$ and a desired control command ${\bm{u}_{(d,t)}}$ by minimizing $J{=}\E[ \norm{ \nd x - \nd x_d}_{\nd Q}^2 + \norm{ \nd u - \nd u_d}_{\nd R}^2]$ In this section, we propose a new closed-loop response model that can explicitly encode the desired states and the control commands by extending the linear feedback controller in SLS to include also a feedforward term as $\nd u = \nd K \nd x + \nd k$. Inserting this into the dynamics equation \eqref{eq:stacked_dynamics}, we obtain $\nd x = \nd Z \nd A_d \nd x + \nd Z \nd B_d(\nd K \nd x + \nd k) + \nd w$, which results in the closed-loop responses
\begin{align}
	\bm{x}&=\Phix\bm{w} + \nd d_{\nd x}, \quad 	\nd d_{\nd x} = \Phix \nd Z \nd B_d \nd k, \label{eq:x_lqt_closed}\\
	\bm{u}&=\Phiu\bm{w} + \nd d_{\nd u}, \quad \nd d_{\nd u} {=} \nd K \nd d_{\nd x} + \nd k, \label{eq:u_lqt_closed}
\end{align} 
where $\Phix$ and $\Phiu$ are defined as in \eqref{eq:resp_x} and \eqref{eq:resp_y}. When we insert $\nd x$ and $\nd u$ in \eqref{eq:x_lqt_closed} and \eqref{eq:u_lqt_closed} into the dynamics equation \eqref{eq:stacked_dynamics}, we get the dynamics constraints on our optimization variables from $\Phix \nd w  + \nd d_{\nd x} {=} \nd S_{\nd x} \nd w + \nd S_{\nd u}(\Phiu \nd w + \nd d_{\nd u})$, which is satisfied for any noise in the system by \eqref{eq:resp_dynamics} and $\nd d_{\nd x} = \nd S_{\nd u} \nd d_{\nd u}$. This can be used to show that $\nd k {=} (\id - \nd K \nd S_{\nd u})\nd d_{\nd u}$. By a few manipulations of equations, one can rewrite the controller into the more interpretable form of $\nd u = \nd K \nd x + \nd k = \nd K (\nd x - \nd d_{\nd x}) + \nd d_{\nd u}$, where $\nd d_{\nd x}$ and $\nd d_{\nd u}$ can be interpreted as the planned trajectory of states and control commands, respectively, assuming zero noise and zero initial state. The feedback part drives the system to a new plan when there is noise, perturbations or nonzero initial state.

The final convex optimization problem becomes
\begin{equation}
	\begin{array}{rl}
		\min_{\Phix,\Phiu, \nd d_{\nd x}, \nd d_{\nd u}}  & \norm{\Phix \nd w + \nd d_{\nd x} - \nd x_d}_{\nd Q}^{2} + 
		\norm{\Phiu \nd w+\nd d_{\nd u}- \nd u_d}_{\nd R}^{2}  \\ 
		\st &  \Phix =\nd S_{x}+\nd S_{u} \Phiu,  \\
		& \nd d_{\nd x} = \nd S_{\nd u} \nd d_{\nd u}, \\
		& \Phix, \Phiu \in \mathcal{L}
	\end{array}
	\label{eq:lqt}
\end{equation}
which can be solved analytically for the optimization variables. For the sake of readability, we omit the details on the derivation and instead give the final result in \Cref{alg:eSLS}.

\begin{algorithm}
	\caption{Extended System Level Synthesis}\label{alg:eSLS}
	\emph{Solve for feedforward terms}
	$\nd d_{\nd u} =( {\nd S_{\nd u}^\trsp}\nd Q\nd S_{\nd u} + \nd R )^{-1}(\nd S_{\nd u}^\trsp\nd Q \nd x_d + \nd R \nd u_d)$\\
	\While( \emph{Solve for feedback terms}){$i<T$}
	{
		\begin{align*}
			\hat{\nd \Phi}^i_u &= -( {\nd S_{\nd u}^{i^\trsp}}\nd Q^i\nd S_{\nd u}^{i} + \nd R^i )^{-1}\nd S_{\nd u}^{i^\trsp}\nd Q^i\nd S_{\nd x}^{i}  \\
			\hat{\nd \Phi}^i_x &= \nd S_{\nd x}^i+\nd S_{\nd u}^i \nd \hat{\nd \Phi}^i_u
		\end{align*}
		
	}
	\emph{Compute the feedback and feedforward parts of the controller with}\\
	 $\bm{K}{=}\Phiu \nd \Phi_x^{-1}$ , $\nd k {=} (\id - \nd K \nd S_{\nd u})\nd d_{\nd u}$
\end{algorithm}


\subsection{Iterative system level synthesis (iSLS)}
\label{subsec:isls}
In this subsection, we consider the problem of system level synthesis for non-linear dynamical systems and non-quadratic cost functions. 
We perform a first order Taylor expansion of the dynamical system $\bm{x}_{t+1}{=}\bm{f}(\bm{x}_t,\bm{u}_t)+\nd w_t$ around some nominal realization of the plant denoted as $(\bm{\hat{x}}_t, \bm{\hat{u}}_t, \nd \mu_{w_t})$, namely, $\bm{x}_{t+1} \approx \bm{f}(\bm{\hat{x}}_t,\bm{\hat{u}}_t) + \nd \mu_{w_t}+  \nd A_t(\deltaxt) + \nd B_t (\deltaut) + (\nd w_t-\nd \mu_{w_t})$ with Jacobian matrices $\{\bm{A}_t \!=\! \frac{\partial\bm{f}}{\partial\bm{x}_t}|_{\bm{\hat{x}}_t, \bm{\hat{u}}_t }, \bm{B}_t \!=\! \frac{\partial\bm{f}}{\partial\bm{u}_t}|_{\bm{\hat{x}}_t, \bm{\hat{u}}_t}\}$. Using the notations $\bm{\hat{x}}_{t+1}{=}\bm{f}(\bm{\hat{x}}_t,\bm{\hat{u}}_t) + \nd \mu_{w_t}$, $\Delta \nd x_t {=} \deltaxt $, $\Delta \nd u_t {=} \deltaut$, the linearized dynamics model can be rewritten as
\begin{align*}
\Delta\bm{x}_{t+1} &= \bm{A}_t \Delta\bm{x}_t + \bm{B}_t \Delta\bm{u}_t+\Delta \nd w_t, \forall t, \\
\iff \Delta \nd x &= \bm{Z}\bm{A}_d\Delta\bm{x} + \bm{Z}\bm{B}_d\Delta\bm{u} + \Delta\bm{w},
\end{align*}
where the second line is the stacked form of the linearized dynamics model. We assume a controller of the form $ \Delta\nd u = \nd K \Delta\nd x + \nd k$ and follow the same procedure described in the previous section to write the closed-loop responses as $\Delta\bm{x}=\Phix\Delta\bm{w}+\nd d_x$ and $\Delta\bm{u}=\Phiu\Delta\bm{w}+\nd d_u$, where the variables $\Phix, \Phiu, \nd d_x, \nd d_u$ satisfy the same constraints in \eqref{eq:lqt} and \eqref{eq:resp_dynamics} with respect to the linearized dynamics model. 

For simplicity, we assume a cost function of the form $c(\bm{x}_t, \bm{u}_t){=}c(\bm{x}_t)+\norm{ \nd u_t}_{\nd R}^2$ where $c(\bm{x}_t)$ is potentially a non-quadratic function of the state. Then, $c(\bm{x}_t)$ can be approximated by a second order Taylor expansion around $\bm{\hat{x}}_t$, namely $c(\bm{x}_t) {\approx} c(\bm{\hat{x}}_t) + \nd c_{\nd x_t}^\trsp(\deltaxt) + \frac{1}{2}(\deltaxt)^\trsp \nd C_{\nd x_t \nd x_t} (\deltaxt) {=}
\frac{1}{2}(\Delta \nd x_t - \nd x_{(d, t)})^\trsp\nd C_{\nd x_t \nd x_t} (\Delta \nd x_t - \nd x_{(d, t)}) + \text{const.}$,
where $\nd c_{\nd x_t}{=}\frac{\partial c}{\partial\bm{x}_t}|_{\bm{\hat{x}}_t, \bm{\hat{u}}_t}$ and $\nd C_{\nd x_t \nd x_t}{=}\frac{\partial^2 c}{\partial\bm{x}_t^2}|_{\bm{\hat{x}}_t, \bm{\hat{u}}_t}$, and $ \nd x_{(d, t)} {=}- \nd C_{\nd x_t \nd x_t}^{-1} \nd c_{\nd x_t}$. This cost can be rewritten in batch form removing the constant terms as $c(\nd x, \nd u) {=} \frac{1}{2}\norm{ \Delta \nd x - \nd x_d}_{\nd C_{\nd x \nd x}}^{2} + \norm{ \Delta \nd u - \nd u_d}_{\nd R}^{2}$.
We propose to perform a Newton's step at each iteration by solving a tracking problem of the same form as \eqref{eq:lqt} by changing the state and the control to $\Delta \nd x$ and $\Delta \nd u$. Thus, this results in the following convex optimization problem
\begin{equation}
	\begin{array}{rl}
		\displaystyle \min_{\Phix,\Phiu,\nd d_x, \nd d_u} & \norm{\Phix \Delta \nd w {+} \nd d_{\nd x} {-} \nd x_d}_{\nd Q}^{2} + 
		\norm{\Phiu \Delta \nd w{+}\nd d_{\nd u}{-} \nd u_d}_{\nd R}^{2}  \\ 
		\st &  \Phix =\nd S_{x}+\nd S_{u} \Phiu,  \\
		& \nd d_{\nd x} = \nd S_{\nd u} \nd d_{\nd u}, \\
		& \Phix, \Phiu \in \mathcal{L}
	\end{array}
	\label{eq:isls}
\end{equation}

which can be solved analytically the same way as eSLS. Newton optimization methods are known to be prone to overshoots, which can be handled via line search. We propose a line search algorithm based on the feedforward control term of our controller as was done by previously proposed iLQR algorithms \cite{ilqr}. Specifically, we define a variable $\alpha $ with a line search strategy with $\nd d_u^{k+1} = \alpha\nd d_u^k$. We accept this update if the trajectories found by these closed loop dynamics models decrease our actual cost function, otherwise we decrease $\alpha$ and re-assess. This corresponds to updating $\nd k$ in the same manner.

\begin{algorithm}[t] 
	\caption{Iterative System Level Synthesis (iSLS)}
	Initialize the nominal state $\bm{\hat{x}}_t$ and control $\bm{\hat{u}}_t$ \;
	Initialize the change in the cost $\Delta c$ \;
	Set a threshold $\tau$ \;
	\While( \emph{Solve iSLS}){$\abs{\Delta c}>\tau$}
	{
		Linearize the dynamics and quadratize the cost function around $\{\bm{\hat{x}}_t,\bm{\hat{u}}_t \}_{t{=}0}^T $  to find $\nd A$, $\nd B$, $\nd C_{\bm{xx}}$, $\nd x_d$ and $\nd u_d$ \;
		Solve \eqref{eq:isls} to find $\nd K$ and $\nd k$ \;
		Do line search to update $\nd k$ using the controller $ \Delta\nd u = \nd K \Delta\nd x + \nd k$ and the dynamics model \;
		Update $\Delta c$.
		
	}
\end{algorithm}

\subsection{Time correlations between states}
\label{subsec:time}
The controller defined by SLS reacts not only to the current state error of the robot, but also to the history of previous states. Such a controller with memory can be exploited in tasks requiring correlations between states at different timesteps, as opposed to a controller without memory such as LQR controllers. An SLS controller can remember what it did in the previous part of the trajectory to use this information to better anticipate the future states and to successfully complete tasks where the future trajectory depends on past states. An example of such task is illustrated in \Cref{fig:setup}.

To achieve correlations between different timesteps, we make use of the off-block-diagonal elements of the precision matrix. Let us denote $\nd x_{t_1}$ and $\nd x_{t_2}$ the states at $t_1$ and $t_2$ that we want to correlate. The correlations that we consider in this work are in the form $\nd C\nd x_{t_1} {+} \nd c{\sim} \nd x_{t_2}$, where $\nd C$ and $\nd c$ are the coefficient matrix and the vector, respectively. We define the correlation cost $c(\nd x_{t_1},\nd x_{t_2})$ as

\begin{align}
	c(\nd x_{t_1},\nd x_{t_2}) &= (\nd C\nd x_{t_1} + \nd c - \nd x_{t_2})^\trsp\bm{Q}_{\text{c}}(\nd C\nd x_{t_1} + \nd c - \nd x_{t_2}),\nonumber \\
	&= \nd x_{t_1}^\trsp\underbrace{\nd C^\trsp\bm{Q}_{\text{c}}\nd C}_{\bm{Q}_{t_1}}\nd x_{t_1} + {(\nd x_{t_2}-\nd c )}^\trsp\underbrace{\bm{Q}_{\text{c}}}_{\nd Q_{t_2}}(\nd x_{t_2}-\nd c )- 2\nd x_{t_1}^\trsp\underbrace{\nd C^\trsp\bm{Q}_{\text{c}}}_{-\bm{Q}_{t_1t_2}}(\nd x_{t_2}-\nd c)  \nonumber \\ 
	& = \matb{\nd x_{t_1} \\ \nd x_{t_2}-\nd c }^\trsp \matb{\bm{Q}_{t_1} && \bm{Q}_{t_1t_2} \\ \bm{Q}_{t_2t_1} && \bm{Q}_{t_2}}\matb{\nd x_{t_1} \\ \nd x_{t_2}-\nd c }.
\end{align}
This means adding off-block-diagonal elements to a typical block-diagonal precision matrix $\nd Q{:=}\mathrm{blkdiag}(\nd Q_0,\nd Q_1, \dots, \nd Q_{T})$, by setting $\nd Q(t_1, t_1) = \nd C^\trsp\bm{Q}_{\text{c}}\nd C$, $\nd Q(t_2, t_2) = \bm{Q}_{\text{c}}$, $\nd Q(t_1, t_2) = -\nd C^\trsp\bm{Q}_{\text{c}}$ and $\nd Q(t_2, t_1) = -\bm{Q}_{\text{c}}\nd C$, where $\nd Q(t_i, t_j)$ represents the precision matrix block corresponding to the timesteps $i$ and $j$.

\subsection{Adaptation to new reference states}
\label{sec:adapt}
We remark that while the system responses encode the information about the precision of the task only, the feedforward part of the responses, namely, $\nd d_{\nd x}$ and $\nd d_{\nd u}$, encode the desired states $\nd x_d$ and the desired control commands $\nd u_d$. Moreover, $\nd d_{\nd x}$ and $\nd d_{\nd u}$ are a linear function of $\nd x_d$ and $\nd u_d$ (see \Cref{alg:eSLS}), which makes them easily recomputable when one changes the desired states and control commands while keeping the precision matrices constant. Examples include trajectory tracking with the same precision throughout the horizon, a task where the robot needs to reach different viapoints and final goal tasks. At the controller level, this only corresponds to a change in the feedforward term $\nd k$, while the feedback part $\nd K$ stays the same. We have $\nd k = (\id - \nd K \nd S_{\nd u})\nd d_{\nd u}, = (\id - \nd K \nd S_{\nd u})( {\nd S_{\nd u}^\trsp}\nd Q\nd S_{\nd u} + \nd R )^{-1}(\nd S_{\nd u}^\trsp\nd Q \nd x_d + \nd R \nd u_d), = \nd F_{\nd x}\nd x_d + \nd F_{\nd u}\nd u_d$, where the terms $\nd F_{\nd x}{=} (\id - \nd K \nd S_{\nd u})( {\nd S_{\nd u}^\trsp}\nd Q\nd S_{\nd u} + \nd R )^{-1}\nd S_{\nd u}^\trsp\nd Q$ and $\nd F_{\nd u}{=} (\id - \nd K \nd S_{\nd u})( {\nd S_{\nd u}^\trsp}\nd Q\nd S_{\nd u} + \nd R )^{-1}\nd R$ can be computed offline and can be used when the desired state and control commands are changed to determine the new feedforward control term $\nd k$ only by matrix-vector multiplication operation along the changed time and dimensions.


In the eSLS formulation with linear dynamics and quadratic cost, the replanned motion and the controller are optimal, i.e., if one wants to optimize from scratch when the desired state is changed, then one ends up with the same solution. However, in iSLS, the replanned motion produces valid trajectories only in the vicinity of the optimal solution. We argue and show that this can still be exploited to produce trajectories which achieve the task successfully without any computation overhead of planning from scratch.

\section{Experiments and Results}
\label{sec:exp}
This section showcases the proposed approach with robotic tasks on a simulated and real environment with Franka Emika robot. The experiment videos for the real task can be found in the video accompanying the paper.

\subsection{Simulated task}
\begin{figure}[t]
	\centering
	\begin{subfigure}{0.45\textwidth}
		\centering
		\includegraphics[width=0.85\textwidth]{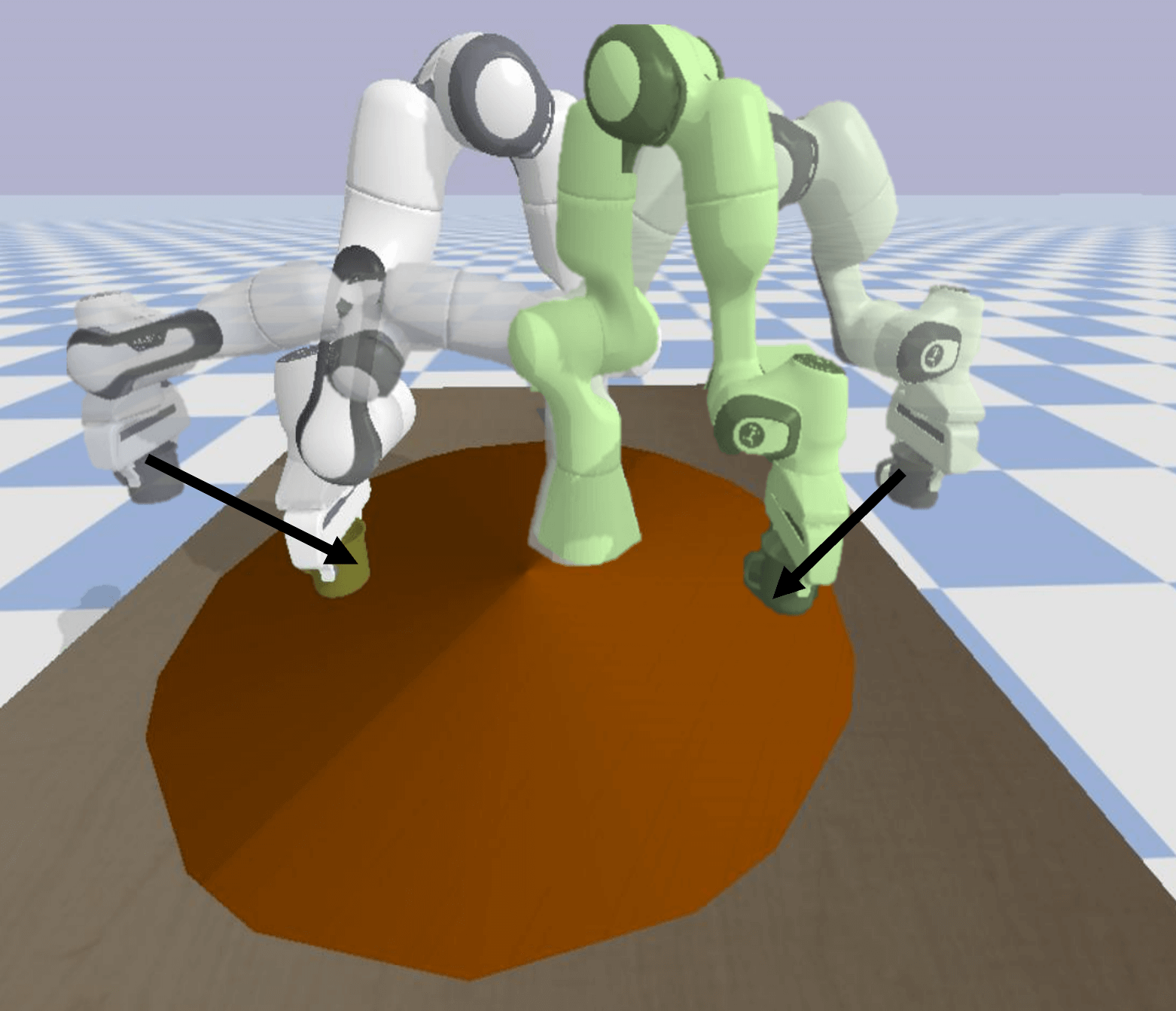}
		\caption{Adaptation to two different initial positions (transparent robots) by following the arrows. }
		\label{fig:sugar_adapt}
	\end{subfigure}\hfill
	\begin{subfigure}{0.45\textwidth}
		\centering
		\includegraphics[width=0.85\textwidth]{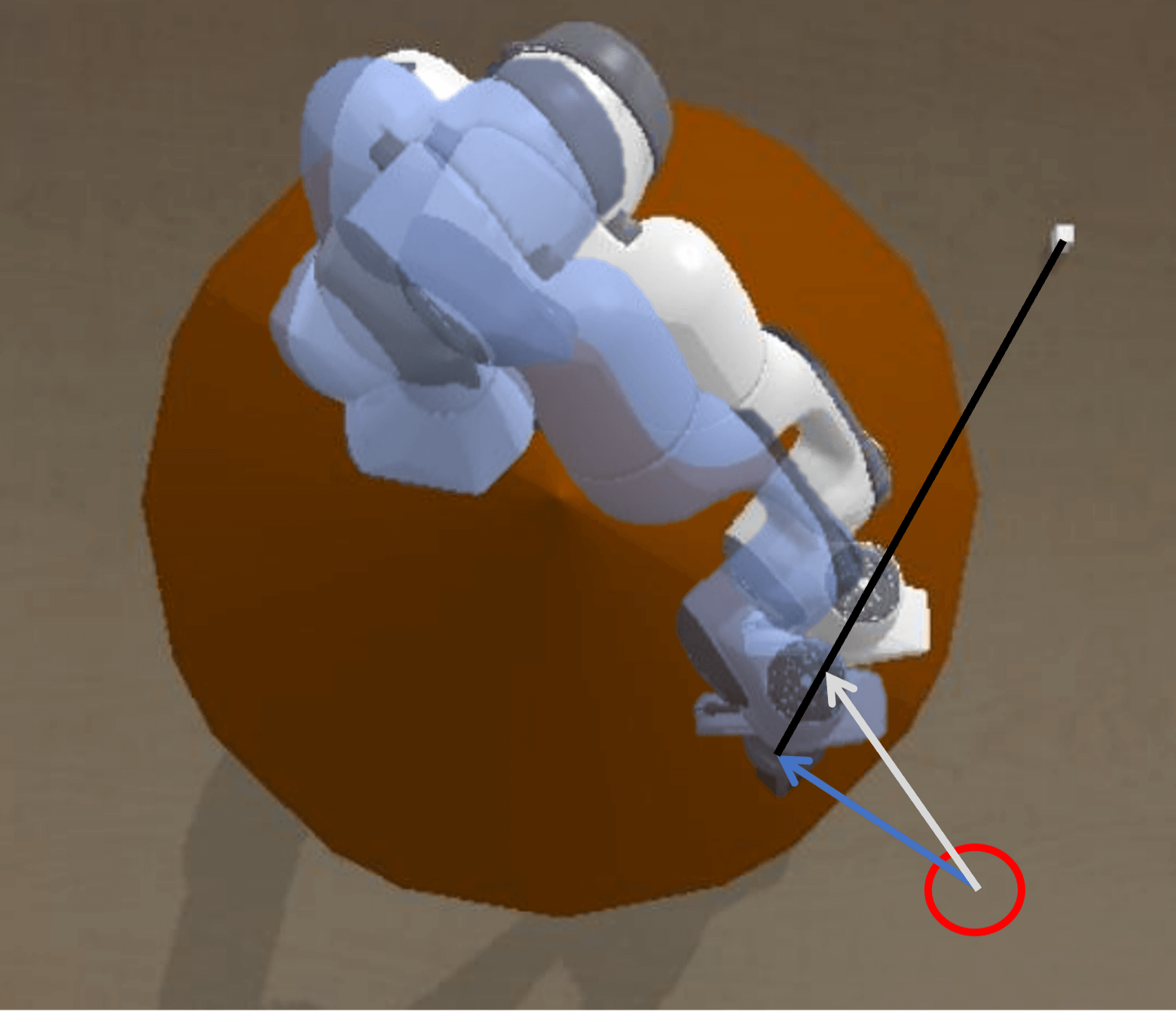}
		\caption{Comparison of the execution of MPC-LQT (blue robot) and eSLS (white robot) controllers}
		\label{fig:comp_qual}
	\end{subfigure}
	\label{fig:sim_task}
	\caption{Simulated task where the robot, initially holding a coffee mug, needs to place it on the table within a certain range (brown disk), pick up a sugar cube (white cube) and drop it onto the mug.}
\end{figure}

We implemented a simulated robotic task exploiting task precisions with the help of the memory feedback. The robot, initially holding a coffee mug, needs to place it on the table within a certain range depicted as brown disk shape on the table (first viapoint, $t{=}20$), pick up a sugar cube depicted as a white cube (second viapoint, $t{=}70$) and drop it onto the mug (goal point, $t{=}100$) as illustrated in \Cref{fig:setup1}. Depending on the initial position of the robot or the perturbations in the environment, the robot can decide on different locations to place the mug. This location needs then to be remembered to drop the sugar cube from the correct location. We implemented eSLS with a double integrator dynamics with 100 timesteps to design a feedback controller on the task space and an inverse kinematics controller to track the reference trajectory generated online by the feedback controller. The cost function has three state cost components and a control cost as $c{=}\norm{\nd x_{20}-\nd c_{d}}_{\nd Q_{20}}^2 + \norm{\nd x_{70}-\nd c_{s}}_{\nd Q_{70}}^2+ \norm{\nd x_{20}-\nd x_{100}}_{\nd Q_{20,100}}^2 + 0.01\norm{\nd u}^2$, where $\nd c_d$ and $\nd c_s$ are the center of the brown disk and the location of the sugar cube, respectively, and $\nd Q_{20}{=}\blkdiag(10^3, 10^3, 10^5, 10^5, 10^5, 10^5)$, $\nd Q_{70}{=}10^5{\times}\id$ $\nd Q_{20,100}{=}\diag(10^5{\times}\id_3, \nd 0_3) $. \Cref{fig:sugar_adapt} shows two examples of eSLS controller starting from two different positions and hence choosing two different locations to place the mug by anticipating that it will need to put the sugar cube inside the mug. These locations differ as there is a trade-off between the state and control costs. As the eSLS controller has memory, the robot can remember this location to accomplish the task.

As a baseline to compare, we chose to design an LQT controller solved by dynamic programming. Obviously, this controller has no information about the previous states and neither about the off-diagonal elements in the precision matrices, hence it can not achieve the task alone. A quick but not generalizable solution to this problem is to recompute the LQT controller after the mug is placed on the table, almost as in MPC, but recomputing the solution only once. We call this controller MPC-LQT. We argue and show that this strategy is far from the optimal behavior because it eliminates the anticipatory and memory aspects of the controller. We tested MPC-LQT against our proposed framework with 10 different initial positions of the robot end-effector, each deciding on different locations of the mug. One such execution is shown in \Cref{fig:comp_qual}, with blue and white robots illustrating MPC-LQT and eSLS strategies, respectively. Each robot, starting from the same initial position depicted by a red circle, places the mug in different locations following the blue and white arrows, and picks up the sugar cube and places it inside the mug correctly. Even though successful, one can see from the geometry in the figure that the final path taken by the MPC-LQT controller (blue) is longer than the proposed eSLS controller (white). Indeed, when we compute the costs of the tasks for both cases, we obtained a cost of 474.3$\pm$204.3 for eSLS and 1004.7$\pm$464.2 for MPC-LQT, which also quantitatively shows the better performance of the behavior produced by our proposed controller. Indeed, after the mug is placed on the table, the recomputed controller of MPC-LQT can also put the sugar cube inside the mug successfully. However, in the optimal behavior as in eSLS, the robot decides on the location of the mug by anticipating that it will need to put the sugar cube inside, which is a missing feature in the baseline.

\begin{figure}[t!]
	\begin{minipage}{\textwidth}
		\centering
		\includegraphics[width=\linewidth]{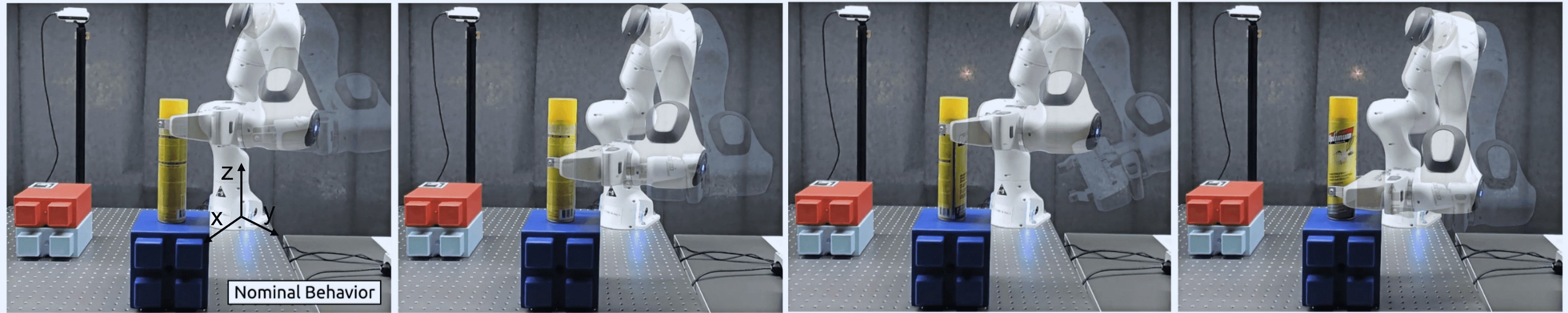}
		\caption{Adaptation of the robot to different initial configurations shown in transparent and the grasping locations shown in solid colors.}
		\label{fig:panda_init_change}
	\end{minipage}
	\begin{minipage}{\textwidth}
		\centering
		\includegraphics[width=\linewidth]{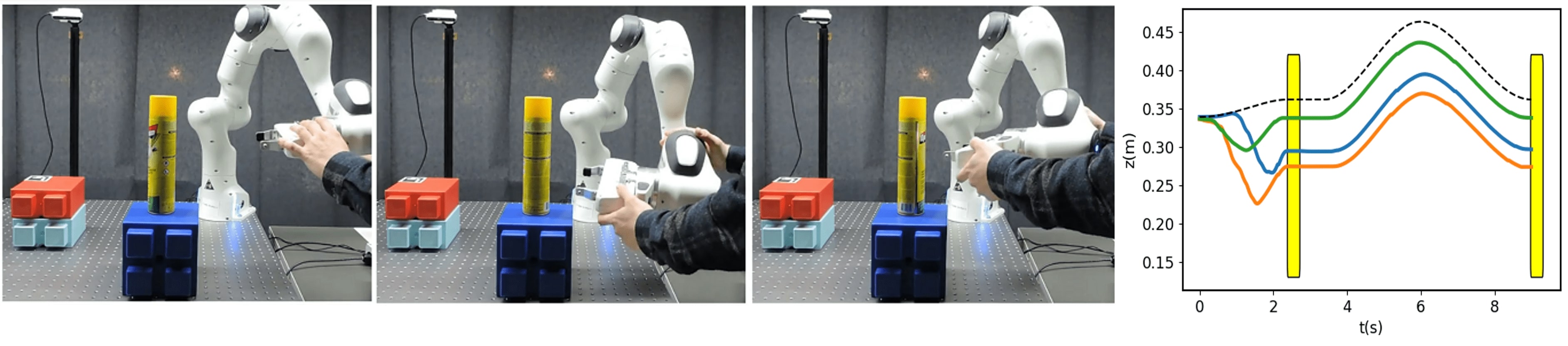}
		\caption{Testing of the reactivity of the controller to different physical perturbations, each resulting in successful completion of the task by adaption. The plot of z-axis position (m) in time (s) gathered from the robot perturbed before grasping the object. The curves of color green, orange and blue correspond to the first, second and third screenshots respectively, while the dashed black line corresponds to the nominal solution. The robot decides on-the-fly to grasp from different locations and remembers these locations to place the yellow object without colliding with the environment.}
		\label{fig:panda_pert}
	\end{minipage}
\end{figure}

\begin{figure}[t]
	\centering
	\includegraphics[width=\textwidth]{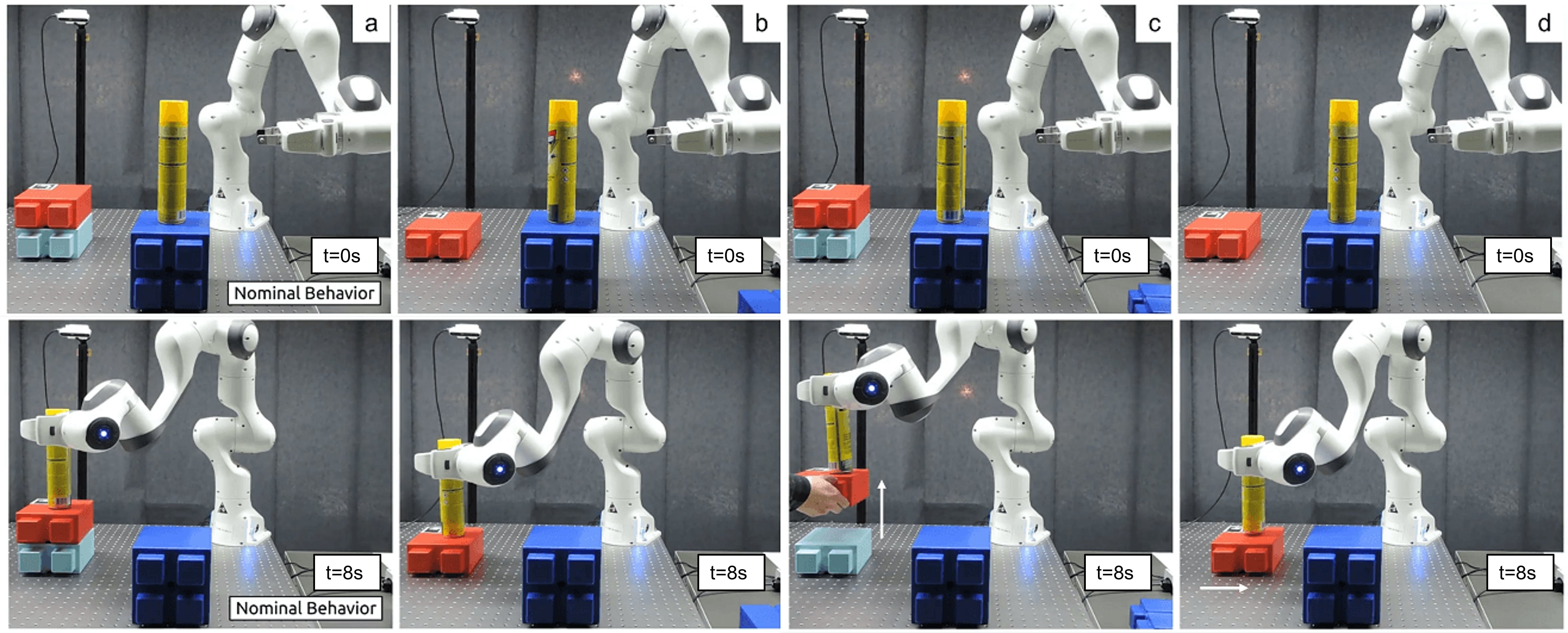}
	\caption{(a) Nominal behavior of the robot without any adaptation. Online adaptation to a change in the final relative height (b) at $t=0s$  (c) at $t=3s$, and (d) at $t=0s$ combined with an on-the-fly change in the final planar position after $t=3s$. The robot adapts to these new goals online with a new plan according to the memory of where it grasped the object.}
	\label{fig:robot_adapt}
\end{figure}

\subsection{Pick-and-place task}
The robot needs to grasp a cylindrical object and place it at a given position while keeping an upright orientation of the end-effector, as seen in \Cref{fig:setup}.  The grasping position is defined by precise x-y positions of the center of the cylinder, whereas the z-position and the rotation around the z-axis are kept at a very low precision. It results in different grasping heights that the robot needs to decide, by anticipating where it is going to place the object. This behaviour exploits the grasping affordances of a cylindrical object by allowing the robot to choose where and how it is going to grasp the object.
The placing location is defined by precise x-y coordinates, while the z-coordinate is defined relatively to the z-coordinate when grasped. Consequently, the robot needs to remember where it grasped the object in order to place it at the relative placing z-position. Notice that without any memory feedback controller, the robot could try to place the object in wrong and dangerous locations as it can force the object to go downwards and collide with the environment.
\subsubsection{Dynamics:}
We denote $\nd \theta$, $\bm{\dot{\theta}}$, $\bm{\ddot{\theta}}$, $\nd x$, $\bm{\dot{x}}$, $\nd q$, the joint angles, the joint velocities, the joint accelerations, the end-effector positions, the end-effector velocities and the end-effector orientation in quaternion format, respectively. We define $\nd f_{\text{kin}}^{\text{pos}}(\cdot)$ and $\nd f_{\text{kin}}^{\text{orn}}(\cdot)$ as the forward kinematics functions outputting end-effector positions and orientations, respectively. We denote $\Log_{\nd q_0}(\nd q_1)$ as the logarithmic map between the quaternions $\nd q_0$ and $\nd q_1$ and define quaternion costs with the methods described in \cite{Calinon20}.

Inspired by the work in \cite{howell2021}, we choose to incorporate the constraints and other nonlinearities into the forward dynamics model of the state in order to alleviate the problems of calculating the Hessian of the cost function. In fact, defining the forward dynamics as such allows to have a quadratic cost function with a precision matrix that defines directly the accuracy of the state itself. This, in turn, becomes very useful when one wants to replan the motion and the controller by changing not only the joint positions, but also the end-effector positions and orientations.
Joint limit constraints can be handled as a soft constraint in the cost function as this approach provides very good results already without resorting to more complicated inequality constrained optimization. However, in this work, we choose to represent these limits inside the state as well to illustrate the generalization and adaptation capability of the controller for a given quadratic cost function. We first define the joint limit violation function as 
$\nd f^{\text{lim}}(\nd \theta) {=} \big(\max{(\nd \theta_l{-}\nd \theta , \nd 0)} {+}\max{(\nd \theta{-}\nd \theta_u , \nd 0)}\big)^2$, where $\nd \theta_l$ and $\nd \theta_u$ are the lower and upper bounds on the joint angles, respectively. Note that when this function is nonzero, it means that the joint limits are violated, which implies that we can use this state as quadratic function to drive it to zero. We then choose to represent the state by $\nd z_t {=}[\nd \theta_t, \bm{\dot{\theta}}_t, \nd x_t, \bm{\dot{x}}_t, {\Log_{\nd q_0}(\nd q_t)}, \nd f^{\text{lim}}] {\in} \mathbb{R}^{31}$ and the control by $\nd u_t {=} \bm{\ddot{\theta}}{\in} \mathbb{R}^{7}$ with the forward dynamics defined as $\matb{\nd \theta_{t+1} &
	\bm{\dot{\theta}}_{t+1}, &
	\nd x_{t+1}, &
	\bm{\dot{x}}_{t+1}, &
	\Log_{\nd q_0}(\nd q_{t+1}), &
	\nd f^{\text{lim}}_{t+1}}\\ =
\matb{
	\nd \theta_{t} {+} \bm{\dot{\theta}}_{t}\delta t, &
	\bm{\dot{\theta}}_{t} {+} \bm{u}_{t}\delta t, &
	\nd f_{\text{kin}}^{\text{pos}}(\bm{\theta}_{t+1}), &
	\nd J(\bm{\theta}_{t+1})\bm{\dot{\theta}}_{t}, &
	\Log_{\nd q_0}(\nd f_{\text{kin}}^{\text{orn}}(\bm{\theta}_{t+1})), &
	\nd f^{\text{lim}}(\bm{\theta}_{t+1})
}$. 

\subsubsection{Cost:}
The cost function is designed by three key points describing the phases of \textit{grasping}, \textit{lifting up} and \textit{placing} at given timesteps. For the \textit{grasping} phase, there is high precision on the x-y axis of the end-effector, and on the end-effector velocity, whereas the precision on the z axis is left very low to give the robot the choice to grasp anywhere along the object. The \textit{lifting} phase is implemented so as to keep the arm safe during the pick-and-place operation to avoid any obstacle on the table. Here, we have high precision only on reaching a z-position 10 cm higher than where the object was grasped during the grasping phase, and no precision on the other dimensions. For the \textit{placing} phase, we have high precision on the x-y positions, on keeping z position the same as where it grasped the object, and on end-effector velocity. Note that this is only possible by exploiting the off-diagonal elements in the precision matrix, as explained in \Cref{subsec:time}. For the orientations, starting from the grasping phase, we set a high precision on keeping an upright orientation (but free to turn around z-axis) in order to keep the balance of the long cylindrical object.

\subsubsection{Implementation:}
We implemented a $50$Hz iSLS controller with a duration of $8$s, which outputs the desired state and control to a $1$kHz impedance controller. A vision system is implemented to track the target object location for testing the adaptation aspect of the controller to the changes in the task. We conducted several experiments with the robot to test the adaptation, reactivity and memory capabilities of the controller. We first tested the adaptation to different initial configurations. We selected randomly 3 configurations that are close to the nominal one, but still corresponding to visibly different end-effector positions and orientations, as seen in \Cref{fig:panda_init_change}. We noticed that the robot could adapt its grasping position to different z-dimensions because of the low precision on that axis. Here, the robot is aware of the possible grasping affordances of the object via our controller and exploits these to decide autonomously from which part of the object to grasp. It then remembers the grasping location and uses it to place the object on the desired position without colliding with the environment.
\subsubsection{Results:}
For testing the reactivity to perturbations of the proposed controller, we applied some force to the robot changing its nominal behavior, i.e. different configurations corresponding to different end-effector positions and orientations, as can be seen in \Cref{fig:panda_pert}. We noticed that the robot reacts to the perturbations by mostly changing its grasping locations, whilst still trying to reach the desired x-y location of the object. Even if it decides only where to grasp the object to cope with the perturbations, it can remember these locations to place the object correctly without collision as can be seen from the robot trajectories in the plot on the far right of \Cref{fig:panda_pert}. In one example shown in the bottom-right of \Cref{fig:panda_pert}, we changed the orientation by turning it around the z-axis. The robot's reaction to this perturbation was considerably smaller than the other perturbation types. This is indeed the expected behavior of the robot, since the error on the orientation with this change stayed very close to zero as a result of the quaternion cost function design.

Finally, we tested the proposed fast adaptation capability of the controller with memory by conducting three different experiments shown by (b), (c) and (d) in \Cref{fig:robot_adapt}. In all cases, we use the same optimized controller which produced the nominal behavior in (a), and reused it to replan in the other cases where the target locations are changed either at the beginning of the movement or on-the-fly. The change in the target location is tracked by the vision system following the marker on the object. In (b), we changed the final height of the placing location to a lower value at the initial time. In (c) and (d), the placing locations are changed on-the-fly by increasing the final height and changing the final x-y position, respectively. We recomputed online the feedforward part of the iSLS controller, namely $\nd k$ as described in \Cref{sec:adapt}, each time there is a significant difference between the current and detected target locations. In all cases, we see that the robot is able to adapt successfully by replanning fast to new desired states without any collision, as can also be seen in the accompanying video.

\section{Conclusion}
\label{sec:conclusion}

We presented an approach for synthesizing reactive and anticipatory controllers that can remember all previous states in the horizon, which is relevant for robotic tasks requiring to have a memory of previous movements. In this context, we proposed to extend SLS controllers to have a feedforward term that can handle viapoint tasks and a Newton optimization method for solving SLS for nonlinear systems and nonquadratic cost functions. We showed that our proposed method outperforms the baseline solutions for producing optimal anticipatory behaviors that require a memory feedback. We showcased our method on a high dimensional robotic system in the presence of perturbations, and demonstrated a step towards adaptation when there is a change of the desired states in the task.

SLS-based representation used in our work can facilitate the bridging between learning, planning and feedback control, especially when we do not know the cost function and/or we do not have prior knowledge on which past states should be correlated with the remaining part of the motion. This provides a very generic formulation for robot skill representations. Future work will study how to determine which past states are correlated by setting the problem as inverse optimal control. We believe that learning such correlations from demonstrations/experiences will be useful to lead the way to the development of smarter feedback controllers which can act on the memory of the states.

SLS offers the advantage that it does not require the experimenter to engineer the problem for each specific use case by augmenting the state-space with the relevant part of the history. In that sense, SLS is generic and formalizes this problem, with a homogeneous formulation, by allowing the system to freely change which past states are correlated and the way they are correlated. In this work, we tackled problems where we see important practical utility of this feature. However, instead of correlating only a couple of timesteps in this work, one could also design these tasks to remember not only one timestep but several timesteps in order to capture an important part of the movement that the robot did in the past and that it needs to remember in the future to react accordingly.
\section*{Acknowledgements}
We would like to thank the reviewers for their thoughtful remarks and suggestions. This work was supported in part by the European Commission’s Horizon 2020 Programme through the CoLLaboratE project (https://collaborate-project.eu/) under grant agreement 820767.
%
%
%

\bibliographystyle{splncs03_unsrt.bst}
\bibliography{bib}

\newpage

\newcommand{\Qd}{\nd Q}
\newcommand{\xd}{\nd \mu_{\nd x}}
\newcommand{\ud}{\nd \mu_{\nd u}}
\newcommand{\inv}{{\raisebox{.2ex}{$\scriptscriptstyle-1$}}}
\newcommand{\dx}{\nd d_{\nd x}}
\newcommand{\du}{\nd d_{\nd u}}

\appendix
	\section{Mathematical background}
	\subsection{Separable costs and constraints}
	Let $\nd A \in \mathbb{R}^{m\times n}$ be a lower triangular matrix, $\nd M \in \mathbb{R}^{m\times m}$ be a matrix and $\nd N \in \mathbb{R}^{n\times n}$ be a diagonal positive definite matrix. Let the superscript $i$ of a matrix denote its column vector (e.g. the vector $\nd A^i \in \mathbb{R}^m$ is the $i^{\text{th}}$ column of $\nd A$) and the superscript $j$ of a matrix denote its row vector (e.g. the vector $\nd A^j \in \mathbb{R}^n$ is the $j^{\text{th}}$ row of $\nd A$). We first observe that  $(\nd A \nd N)^i{=}\nd A \nd N^{i}{=}\nd A^i \nd N^{ii}$ as $\nd N$ is a diagonal matrix, and $(\nd M \nd A)^i{=}\nd M \nd A^{i}{=}\nd M^{i:}\nd A^{i}$ as $\nd A$ is a lower triangular matrix, where $\nd M^{i:}$ is the submatrix obtained by removing all the rows and columns before the index $i$. Then, we can obtain the following identities: 
	\begin{align}
		\norm{\nd A}_{F}^2&=\sum_i \sum_j A_{ij} = \sum_i \norm{\nd A^i}_2^2 = \sum_j \norm{\nd A^j}_2^2,\\
		\norm{\nd A\nd N}_{F}^2&= \sum_i \norm{(\nd A\nd N)^i}_2^2=\sum_i \norm{\nd A^i\nd N^{ii}}_2^2, \\
		\norm{\nd M \nd A}_{F}^2&= \sum_i \norm{(\nd M\nd A)^i}_2^2=\sum_i \norm{\nd M^{i:}\nd A^{i}}_2^2, \\
		\iff &\norm{\nd M \nd A \nd N}_{F}^2= \sum_i \norm{\nd M^{i:}\nd A^{i}\nd N^{ii}}_2^2
	\end{align} 
	Let $\nd P\in \mathbb{R}^{n\times n}$, $\nd B\in \mathbb{R}^{n\times n}$ and $\nd R\in \mathbb{R}^{n\times m}$ be lower triangular matrices. Then, the equality $\nd B{=}\nd P + \nd R \nd A$ can be column-wise separated as $\nd B^i{=}\nd P^{i} + \nd R^{i:} \nd A^i$

	\subsection{Expectation of some linear and quadratic forms}
	Let $\nd x{\sim}\normal(\nd \mu,\nd \Sigma)$, then
	\begin{align}
		\E_{\nd x}[\nd A \nd x + \nd a] &= \nd A \nd \mu + \nd a, \\
		\E_{\nd x}[\norm{\nd A \nd x + \nd a}_2^2] &= \norm{\nd A \nd \Sigma^{\frac{1}{2}}}_F^2 + \norm{\nd a}_2^2,\\
		\E_{\nd x}[\norm{\nd A \nd x + \nd a}_{\nd Q}^2] &= \norm{\nd A \nd \Sigma^{\frac{1}{2}}}_{\nd Q}^2 + \norm{\nd a}_{\nd Q}^2, \\
		\E_{\nd x}[(\nd A \nd x + \nd a)^\trsp(\nd B \nd x + \nd b)] &= \trace{\nd A \nd \Sigma \nd B} + \nd a^\trsp \nd b
	\end{align}
	
	\section{Linear quadratic tracking with least-squares}
	A linear quadratic problem as in
	\begin{equation}
		\begin{array}{rl}
			\min_{\nd x,\nd u} & \norm{ \nd x - \xd}_{\Qd}^2 + \norm{ \nd u}_{\nd R}^2\\
			\st & \nd x = \Sx \nd w + \Su \nd u,
		\end{array}
	\end{equation}
	can be solved for $\nd u$ analytically by replacing $\nd x$ in the cost function by the equality constraint and transforming the cost function to be indepedent of $\nd x$ as $J_{\text{batch-lqt}}{=}\norm{ \Sx\nd w + \Su \nd u - \nd x_d}_{\nd Q}^2 + \norm{ \nd u - \nd u_d}_{\nd R}^2$. This is a quadratic function of $\nd u$ and can be solved for $\nd u$ in the least-squares sense as in
	\begin{equation*}
		\bm{\hat{u}} = (\Su^\trsp \Qd \Su + \nd R)^\inv\Su^\trsp \Qd (\xd -\Sx \nd x_0) 
	\end{equation*}
		
	\section{SLS controller derivation}
	For $\nd w{\sim}\normal(\nd 0,\nd \Sigma_{w})$, where $\Sigma_{w}$ is a diagonal matrix with diagonal entries denoted as $\sigma^i$, taking the expectation of the SLS cost, we obtain
	\begin{align}
		J_{\text{SLS}}&=\E_{\nd w}[\norm{\nd \Phi_x \nd w }_{\nd Q}^{2} +  \norm{\nd \Phi_u \nd w}_{\nd R}^{2}], \nonumber \\
		&= \norm{\nd \Phi_x \nd \Sigma_{w}^{\frac{1}{2}} }_{\nd Q}^{2} +  \norm{\nd \Phi_u \nd \Sigma_{w}^{\frac{1}{2}}}_{\nd R}^{2}, \\
		&= \sum_{i=1}^T\underbrace{\norm{\nd \Phi_x^i \nd \sigma_{w}^{i} }_{\nd Q^{i:}}^{2} +  \norm{\nd \Phi_u^i \nd \sigma_{w}^{i}}_{\nd R^{i:}}^{2}}_{J_{\text{SLS}}^i(\nd \Phix^i, \nd \Phiu^i)},
	\end{align}
	which shows that the cost function can be separated into $T$ terms each ($i$) depending on only on the $i^{\text{th}}$ block-column of $\nd \Phi_x$ and $\nd \Phi_u$. We can separate column-wise also the dynamics constraint $\Phix = \Sx + \Su \Phiu$ as in $\Phix^i = \Sx^i + \Su^{i:}\Phiu^i$ and set up the SLS optimization subproblem $i$ as 
	\begin{equation}
		\begin{array}{rl}
			\displaystyle \min_{\Phix^i,\Phiu^i} & \sum_{i=1}^T J_{\text{SLS}}^i(\nd \Phix^i, \nd \Phiu^i) \\ 
			\st &  \Phix^i = \Sx^i + \Su^{i:}\Phiu^i,  \\
		\end{array}
		\label{eq:sls_i}
	\end{equation}
	The solution to \eqref{eq:sls_i} for each $i \in [1,\cdots, T]$ can be found analytically. By inserting the linear constraint inside the cost function, we convert \eqref{eq:sls_i} to a least-square problem by $	\Phiu^{i^*}=\argmin_{\Phiu^i}J_{\text{SLS}}^i(\Phiu^i)=\norm{(\Sx^i + \Su^{i:}\Phiu^i) \nd \sigma_{w}^{i} }_{\nd Q^{i:}}^{2} +  \norm{\Phi_u^i \nd \sigma_{w}^{i}}_{\nd R^{i:}}^{2} $ which admits the following analytical solution:
	\begin{align}
	\Phiu^{i^*}&= -( \Su^{{i:}^\trsp}\nd Q^{i:}\Su^{i:} + \nd R^{i:} )^\inv\Su^{{i:}^\trsp}\nd Q^{i:}\Sx^i, \\
	\Phix^{i^*} &= \Sx^i+\Su^{i:} \Phiu^{i^*}.
	\label{eq:sls_sol_i}
	\end{align}
	The eSLS cost can be expressed as
	\begin{align}
		J_{\text{eSLS}}&=\E_{\nd w}[\norm{\Phix \nd w + \dx - \xd}_{\nd Q}^{2} +  \norm{\Phiu \nd w + \du- \ud}_{\nd R}^{2}], \nonumber \\
		&= \norm{\nd \Phi_x \nd \Sigma_{w}^{\frac{1}{2} }}_{\nd Q}^{2} + \norm{\dx - \xd}_{\nd Q}^2 + \norm{\nd \Phi_u \nd \Sigma_{w}^{\frac{1}{2}}}_{\nd R}^{2} + \norm{\du - \ud}_{\nd R}^2, \nonumber \\
		&= J(\dx, \du) + \sum_{i=1}^T J_{\text{SLS}}^i(\nd \Phix^i, \nd \Phiu^i)
		\label{eq:esls_cost}
	\end{align}
	which shows that the cost function can be decomposed into a part $J(\dx, \du)=\norm{\dx - \xd}_{\nd Q}^2+\norm{\du - \ud}_{\nd R}^2$ that optimizes separately the feedforward pair $\{\dx, \du\}$ and a part $J_{\text{SLS}}^i(\nd \Phix^i, \nd \Phiu^i)$ that optimizes the feedback pair $\{\Phix^i, \Phiu^i\}$. The solution to the feedback pair is given by \eqref{eq:sls_sol_i} and the solution to the feedforward pair can be determined solving the following optimization problem: 
		\begin{equation}
		\begin{array}{rl}
			\displaystyle \min_{\dx,\du} & J(\dx, \du) \\ 
			\st &  \dx = \Su\du  \\
		\end{array}
		\label{eq:esls_ff}
	\end{equation}
	The solution to \eqref{eq:esls_ff} can be found analytically. By inserting the linear equality constraint inside the cost function, we convert \eqref{eq:esls_ff} to a least-square problem by $\du^*=\argmin_{\du}\norm{\Su\du - \xd}_{\nd Q}^2+\norm{\du - \ud}_{\nd R}^2$ which admits the following analytical solution:
	\begin{align}
		\du^* &= ( \Su^\trsp\nd Q\Su + \nd R )^\inv(\Su^\trsp\nd Q \xd + \nd R \ud), \\
		\dx^* &= \Su \du^*
		\label{eq:esls_ff_sol}
	\end{align}

\end{document}